\documentclass[sigconf]{acmart}
\AtBeginDocument{%
  }

\setcopyright{acmlicensed}
\copyrightyear{2026}
\acmYear{2026}
\acmDOI{XXXXXXX.XXXXXXX}
\acmConference[Conference acronym 'XX]{Make sure to enter the correct
  conference title from your rights confirmation email}{June 03--05,
  2018}{Woodstock, NY}
\acmISBN{978-1-4503-XXXX-X/2018/06}

\usepackage{enumitem}
\usepackage{xspace}
\usepackage{url}
\usepackage{balance}
\usepackage{amsmath}
\usepackage{amsfonts}
\usepackage{multirow}
\usepackage{booktabs}
\usepackage{algorithmic}
\usepackage{graphicx}
\usepackage{textcomp}
\usepackage{xcolor}

\usepackage{booktabs}

\newcommand{\stitle}[1]{\noindent {\bf {#1}}}
\newcommand{\sstitle}[1]{\noindent{\underline{\textit {#1}}}}

\newcommand{\kw}[1]{{\ensuremath {\mathsf{#1}}}\xspace}
\newcommand{\bkw}[1]{{\ensuremath {\mathsf{\textbf{#1}}}}\xspace}
\newcommand{\euler}{\kw{EulerESG}}
\newcommand{\beuler}{\bkw{EulerESG}}

\begin{document}

\title{EulerESG: Automating ESG Disclosure Analysis with LLMs}

\author{Yi Ding}
\affiliation{%
  \institution{UNSW Sydney}
  \country{} % \country{Australia}
}
\email{yi.k.ding@unsw.edu.au}

\author{Xushuo Tang}
\affiliation{%
  \institution{UNSW Sydney}
  \country{}
}
\email{xushuo.tang@unsw.edu.au}

\author{Zhengyi	Yang}\thanks{* Zhengyi Yang is the corresponding author.}
\affiliation{%
  \institution{UNSW Sydney}
  \country{}
}
\email{zhengyi.yang@unsw.edu.au}

\author{Wenqian Zhang}
\affiliation{%
  \institution{UNSW Sydney}
  \country{}
}
\email{wenqian.zhang@unsw.edu.au}

\author{Simin Wu}
\affiliation{%
  \institution{Eigenflow AI}
  \country{}
}
\email{sm.wu@eigenflow.ai }

\author{Yuxin Huang}
\affiliation{%
  \institution{Euler AI}
  \country{}
}
\email{yuxin.huang@eulerai.au}

\author{Lingjing Lan}
\affiliation{%
  \institution{UNSW Sydney}
  \country{}
}
\email{lingjing.lan@unsw.edu.au}

\author{Weiyuan Li}
\affiliation{%
  \institution{UNSW Sydney}
  \country{}
}
\email{felix.li2@unsw.edu.au}

\author{Yin	Chen}
\affiliation{%
  \institution{UTS Sydney}
  \country{}
}
\email{yin.chen@student.uts.edu.au}

\author{Mingchen Ju}
\affiliation{%
  \institution{UNSW Sydney}
  \country{}
}
\email{mingchen.ju@unsw.edu.au}

\author{Wenke Yang}
\affiliation{%
  \institution{UNSW Sydney}
  \country{}
}
\email{wenke.yang@unsw.edu.au}

\author{Thong Hoang}
\affiliation{%
  \institution{Data61, CSIRO}
  \country{}
}
\email{james.hoang@data61.csiro.au}

\author{Mykhailo Klymenko}
\affiliation{%
  \institution{Data61, CSIRO}
  \country{}
}
\email{mike.klymenko@data61.}
\email{csiro.au}

\author{Xiwei Xu}
\affiliation{%
  \institution{Data61, CSIRO}
  \country{}
}
\email{xiwei.xu@data61.csiro.au}

\author{Wenjie Zhang}
\affiliation{%
  \institution{UNSW Sydney}
  \country{}
}
\email{wenjie.zhang@unsw.edu.au}

\renewcommand{\shortauthors}{Ding et al.}

\begin{abstract}
Environmental, Social, and Governance (ESG) reports have become central to how companies communicate climate risk, social impact, and governance practices, yet they are still published primarily as long, heterogeneous PDF documents. This makes it difficult to systematically answer seemingly simple questions. Existing tools either rely on brittle rule-based extraction or treat ESG reports as generic text, without explicitly modelling the underlying reporting standards. We present \textbf{EulerESG}, an LLM-powered system for automating ESG disclosure analysis with explicit awareness of ESG frameworks. EulerESG combines (i) dual-channel retrieval and LLM-driven disclosure analysis over ESG reports, and (ii) an interactive dashboard and chatbot for exploration, benchmarking, and explanation. Using four globally recognised companies and twelve SASB sub-industries, we show that EulerESG can automatically populate standard-aligned metric tables with high fidelity (up to 0.95 average accuracy) while remaining practical in end-to-end runtime, and we compare several recent LLM models in this setting. The full implementation, together with a demonstration video, is publicly available at \url{https://github.com/UNSW-database/EulerESG}.
\end{abstract}


\keywords{ESG, Sustainability, LLM, AI Agent}

\maketitle

\section{Introduction}
\label{sec:intro}
Environmental, Social, and Governance (ESG) refers to non-financial criteria for evaluating corporate sustainability and ethical conduct. \textit{Environmental} factors address climate impact, energy use, and waste; \textit{social} factors include labor rights, diversity, and community engagement; and \textit{governance} concerns leadership, compliance, and transparency. ESG has become a key consideration for stakeholders assessing long-term corporate viability. Studies show that strong ESG performance correlates with lower risk, greater operational efficiency, lower capital costs, enhanced brand reputation, improved operational performance, and higher employee retention\cite{friede2015esg,khan2016corporate,whelan2021esg,shakil2021environmental}. ESG practices also align with the United Nations Sustainable Development Goals (SDGs), providing various reporting frameworks, such as the Sustainability Accounting Standards Board (SASB) and the Global Reporting Initiative (GRI). 

% ESG disclosures have recently played a central role in promoting corporate accountability and sustainability~\cite{whelan2021esg,shakil2021environmental,chauke2018three}. Regulatory frameworks worldwide are increasingly mandating such disclosures. In the European Union, the Corporate Sustainability Reporting Directive (CSRD) requires large companies to publish detailed ESG reports, covering approximately $50{,}000$ entities~\cite{eu_csrd}. Australia will enforce mandatory climate-related disclosures for major businesses starting in 2025~\cite{australia_mandatory}. China is progressing toward ISSB-aligned mandatory ESG reporting by 2030~\cite{china_esg}. In the United States, the Securities and Exchange Commission (SEC) adopted a 2024 rule requiring listed firms to disclose climate risks, greenhouse gas emissions, and governance structures~\cite{us_sec}. These developments underscore the global momentum toward standardized and transparent ESG reporting.

% At the global level, the United Nations Global Compact (UNGC) encourages over 20,000 participating companies to align their ESG practices with principles on human rights, labor, environment, and anti-corruption~\cite{ungc_2023}.

To meet growing investor expectations and regulatory requirements, organizations publish ESG disclosures to document their sustainability efforts and associated risks. These disclosures typically appear in stand-alone ESG reports.
ESG reports serve multiple purposes: supporting compliance checks, enhancing transparency, aiding investor decision-making, enabling benchmarking, and aligning with voluntary or mandatory frameworks such as GRI~\cite{gri}, SASB~\cite{sasb} and TCFD~\cite{tcfd}. However, despite the increasing adoption of these standards, ESG disclosures often have complex layouts, are highly variable across sectors, and are predominantly published in PDF format, making them difficult for automated processing and analysis. These reports typically contain both structured metrics (e.g., greenhouse gas emissions, employee turnover) and unstructured narratives (e.g., board diversity policies, climate resilience strategies).
As a result, large-scale ESG analysis remains labor-intensive, requiring extensive manual review, significantly limits its utility in automated compliance checks and decision-support systems.
We aim to develop an automated tool that efficiently and effectively supports the analysis of ESG reports.

% Beyond compliance, developing and disclosing ESG practices brings tangible benefits to corporations. Research shows that companies with strong ESG performance often enjoy lower capital costs, enhanced brand reputation, improved operational efficiency, and higher employee retention~\cite{friede2015esg, khan2016corporate}. ESG disclosure also helps firms anticipate regulatory changes and societal expectations, reducing long-term risks.

% From a societal perspective, transparent ESG reporting contributes to broader sustainability goals by enabling data-driven environmental and social interventions~\cite{gibassier2020esg}. Publicly disclosed ESG data empowers stakeholders—including regulators, civil society, and consumers—to hold companies accountable and fosters responsible business conduct that aligns with the UN Sustainable Development Goals (SDGs)~\cite{ungc_2023}.

\stitle{Challenges.}  
Most ESG information is embedded within lengthy, unstructured PDF reports, which vary widely in layout, terminology, and reporting practices. This heterogeneity poses significant challenges for automated analysis~\cite{peng2024advanced, wang2023scalable}.
The key challenges in automating ESG report analysis are:

\sstitle{Unstructured and Heterogeneous Formats.} 
    ESG content appears in diverse layouts and structures, making reliable extraction of key metrics difficult for rule-based systems.
    
    % \item \sstitle{Inconsistent Terminology and Granularity.} 
    % Companies adopt different terms, units, and levels of detail, even within the same sector. This inconsistency complicates the identification and comparison of metrics.

\sstitle{Inconsistent Terminology and Irrelevant Content.}  
    Companies often use varying terms, units, and levels of detail to describe similar metrics. Additionally, ESG reports frequently contain excessive promotional content that does not contribute to substantive disclosures which hinder the accurate comparison of relevant ESG metrics.

\sstitle{Cross-Industry and Multi-Standard Alignment.}  
    Aligning disclosures with ESG frameworks is challenging due to mismatches in scope, taxonomy, and expectations across standards and industries.

% \textit{(We address this with a robust PDF parsing pipeline tailored for ESG contexts.)}
% \textit{(We tackle this through prompt-optimized LLMs that recognize ESG entities and normalize expressions contextually.)}
% \textit{(We build a dynamic ESG standard library to support cross-framework and cross-industry alignment.)}

% \item \textbf{Scattered Information Across Text and Tables}:
% ESG metrics often appear in fragmented formats—buried within tables, figures, or long narratives—requiring semantic understanding and contextual integration.
% \textit{(We apply LLM-based entity recognition techniques to extract and classify relevant information across modalities.)}

% \item \textbf{Limitations of Existing NLP Tools}:
% Traditional NLP models struggle with long documents, domain-specific jargon, and document-level context aggregation, especially when handling industry-specific ESG disclosures.
% \textit{(We leverage prompt engineering and domain adaptation to fine-tune LLM behavior for ESG-specific tasks.)}

\stitle{Contributions.}  
With the recent advances of Large Language Model (LLM)~\cite{zhang2024mm,openai2023gpt4,rawat2024recent,li2024citation}, we present \beuler, a LLM-powered tool for interactive ESG report analysis.
%
% \euler extracts structured ESG metrics from unstructured corporate reports, maps them to standardised frameworks, and identifies missing or partial disclosures. 
%
Our contributions are as follows:

\sstitle{ESG-Standard-Specific Metrics Extraction.} 
  We design an automated pipeline that leverages LLMs to extract structured ESG metrics from unstructured, noisy, and highly variable corporate reports in PDF format, significantly reducing the need for manual rule engineering.

\sstitle{Multi-Framework Standard Alignment.}  
  We cover over 100 industries and multiple reporting frameworks. Our mapping module enables fine-grained alignment of extracted disclosures with industry-specific requirements, supporting flexible comparisons across companies, industries, and regulatory standards.

\sstitle{LLM-Powered Automated ESG Analysis.} 
  We design an ESG agent that integrates LLM reasoning with prompt engineering to perform context-aware entity recognition, metric classification, and compliance flagging. The agent enables benchmarking and decision support through interactive interfaces, allowing users to query ESG reports in natural language and receive contextualized insights.

% \stitle{Outline.} We first provide a brief, self-contained overview of the
% ESG landscape and its data-management challenges
% (Section~\ref{sec:esg-background}). We focus in particular on how
% regulatory frameworks such as SASB define a structured space of
% industries, sub-industries, and metrics, and why this structure is
% crucial for building standard-aware, machine-readable ESG datasets.
% Section~\ref{sec:system} then introduces the design of \euler,
% followed by our experimental evaluation in
% Section~\ref{sec:experiments}.

% To demonstrate the feasibility and performance of \euler, we implement an interactive web-based system that allows users to upload ESG reports, automatically extract relevant disclosures, visualize them in structured tables, and highlight coverage gaps.

% \paragraph{Outline.}The rest of this paper is organized as follows: Section~\ref{sec:background} provides background on ESG reporting and related work. Section~\ref{sec:system} details the architecture and implementation of the \euler system. Section~\ref{sec:demo} presents a user-facing demonstration with real-world use cases across industries. Section~\ref{sec:conclusion} concludes with future directions and planned enhancements.

% \input{sec2-bg}
\vspace{-0.8cm}
\section{System Design}
\label{sec:system}

 \begin{figure*}[t]
  \centering
  \includegraphics[width=0.9\textwidth]{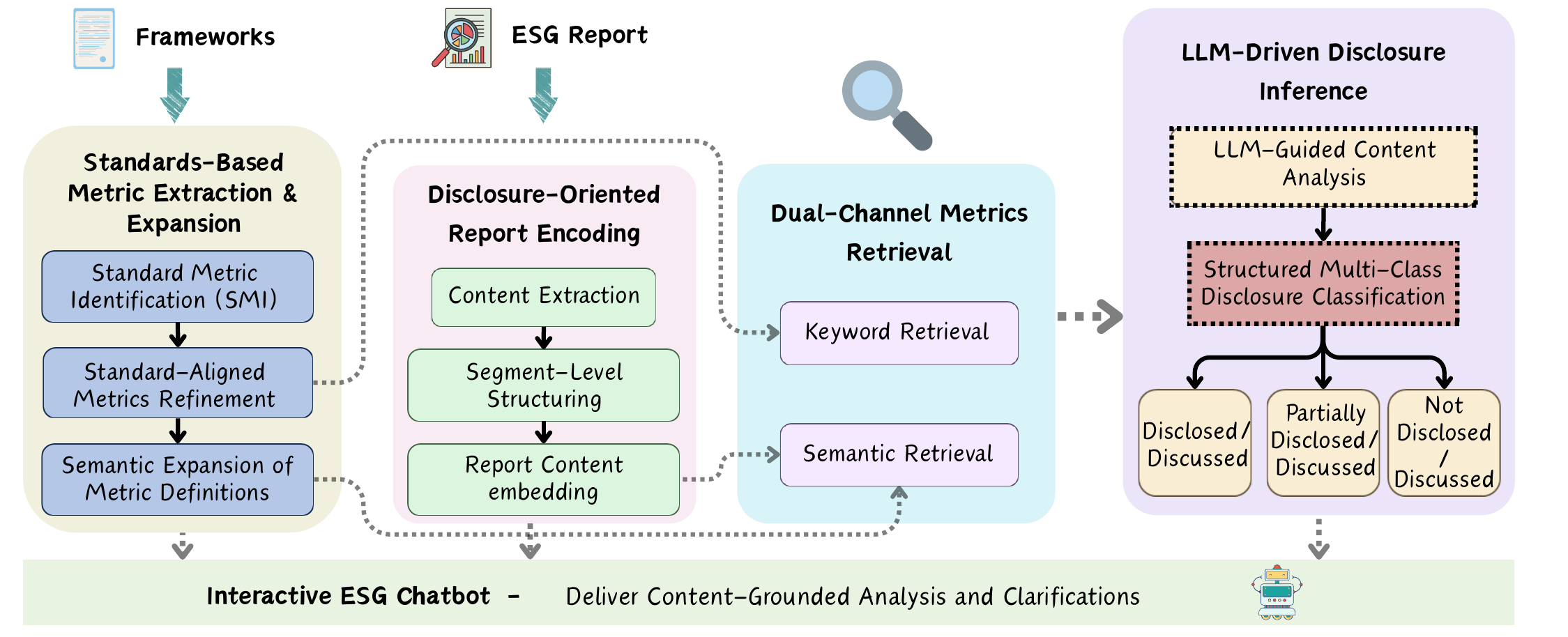}
  \vspace{-0.2cm}
  \caption{\beuler Workflow.}
  \label{fig:workflow}
  \vspace{-0.5cm}
\end{figure*}

% \subsection{Workflow}
Figure \ref{fig:workflow} shows the five-module design of \euler. The architecture reflects a deliberate design choice to decompose the complex ESG analysis task into manageable, specialized processing stages while maintaining a systematic information flow throughout the evaluation process.

\sstitle{Standards-Based Metric Extraction \& Expansion.}   
This module preprocesses ESG standard metrics through a three-stage refinement process. 
First, the \textit{Standard Metric Identification (SMI)} component extracts relevant metrics from an extensive collection of unstructured ESG standards. 
Second, we apply \textit{Standard-Aligned Metrics Refinement}, which augments each metric with additional metadata and descriptions derived from ESG frameworks. 
Finally, we introduce the \textit{Semantic Expansion of Metric Definitions}, which enriches each metric with contextual information necessary for vector-based retrieval. This expansion supports semantic matching between standard definitions and ESG report content, enhancing the system’s ability to identify relevant disclosures.

\sstitle{Disclosure-Oriented Report Encoding.}  
Analyzing ESG reports poses significant challenges due to their heterogeneous formats. This module addresses these challenges through a systematic three-stage approach.
First, \textit{Content Extraction} is performed page by page. When a paragraph or table spans multiple pages, we retain the full structure to preserve contextual integrity.
Next, \textit{Segment-Level Structuring} annotates the extracted content at the segment level (e.g., by paragraph), enabling precise positioning of metrics for downstream analysis.
Finally, \textit{Report Content Embedding} transforms each textual segment into a vector representation using an embedding model. In our implementation, we adopt the open-source BGE-M3 model~\cite{chen2024bge}. This creates a searchable content foundation that supports the \textit{Dual-Channel Metrics Retrieval Modules} used in the subsequent module.

\sstitle{Dual-Channel Metrics Retrieval.}  
When retrieving information from ESG reports, we adopt a \textit{Dual-Channel Metrics Retrieval Module} that combines \textit{Keyword Retrieval} and \textit{Semantic Retrieval}.
Keyword retrieval searches directly for occurrences of metric terminology within the report. 
%
% For example, using SASB standards, each company is classified into its corresponding industry category, determining the relevant metric set for that industry. 
%
However, due to the variation in language and structure across ESG frameworks and corporate reports, exact keyword matches are often insufficient.
To address this, we implement a complementary Semantic Retrieval mechanism, which uses the embeddings generated during the semantic expansion phase to identify relevant segments in the report. Specifically, we apply vector similarity matching between extended metric definitions and report content using BGE-M3 embeddings, then re-rank all candidates with the BGE-Reranker-v2-M3 model~\cite{baai2024bgereranker} to refine relevance scores. When both channels retrieve the same segment, we combine their scores with higher weight assigned to reranker results. The top-5 highest-scoring segments are retained for subsequent LLM-based disclosure analysis.
%
% This dual-retrieval design ensures both high recall and precision, enabling comprehensive and reliable content identification for analysis.

\sstitle{LLM-Driven Disclosure Analysis.}  
This module transforms retrieved content into systematic compliance assessments through a two-phase reasoning process.
\textit{LLM-Guided Content Reasoning} employs designed analytical prompts to examine retrieved content segments, evaluating both the indicators and the quality of disclosures based on structured criteria.
Then, the results undergo \textit{Structured Multi-Class Disclosure Classification}, categorizing each result into one of three predefined classes: \stitle{(1) (Fully) Discussed/Disclosed, }
% : The metric is clearly reported, supported by quantitative values or detailed discussed. 
\stitle{(2) Partially Discussed/Disclosed, }
% : The topic is mentioned, but lacks concrete values or sufficient detail. 
\stitle{(3) Not Discussed/Disclosed. }
% : The metric is entirly absent from the report. 
Given common ESG frameworks, a metric is typically classified as either (1) quantitatively \textbf{Disclosed}, where a measurable value is expected; or (2) qualitatively \textbf{Discussed}, involving narrative context such as policies, risks, or governance structures.

% The framework concludes with capabilities that address the critical need for analytical transparency and stakeholder engagement in compliance assessment outcomes. We recognize that automated compliance evaluation must remain accountable and interpretable to maintain credibility and support decision-making processes. Through content-grounded explanations and conversational interfaces, users can explore the analytical foundations underlying each assessment decision, access detailed reasoning traces, and understand the connection between source evidence and evaluation outcomes.
\sstitle{Interactive LLM ESG Chatbot.}  
The integrated LLM-powered chatbot allows users to interact with ESG reports through natural language queries, enabling more user-friendly and accessible exploration.
For example, the chatbot supports:  
(1) on-demand definitions of ESG terms and metric codes;  
(2) contextual queries about metrics or policies not explicitly covered by standard frameworks;  
(3) retrieval of values or discussion locations from the original PDF; and  
(4) summarization of policy narratives or ambiguous disclosures.
%
% The chatbot uses the GPT-5 model~\cite{hurst2024gpt}. 
%
The LLM connects to vectorized content generated by other modules, providing access to knowledge about ESG standards, ESG reports, and analysis results. This integration enables the chatbot to answer domain-specific questions accurately and contextually.

% We use BGE-M3 to vectorized the output of each module, and then use BGE-Rerank-v2-M3 to retrieve the vectorized content based on user questions, enhancing the chatbot's performance in answering open-ended metric questions of interest to users and improving the interpretability of the entire system.

\begin{figure*}[ht]
  \centering
  \includegraphics[width=0.32\textwidth]{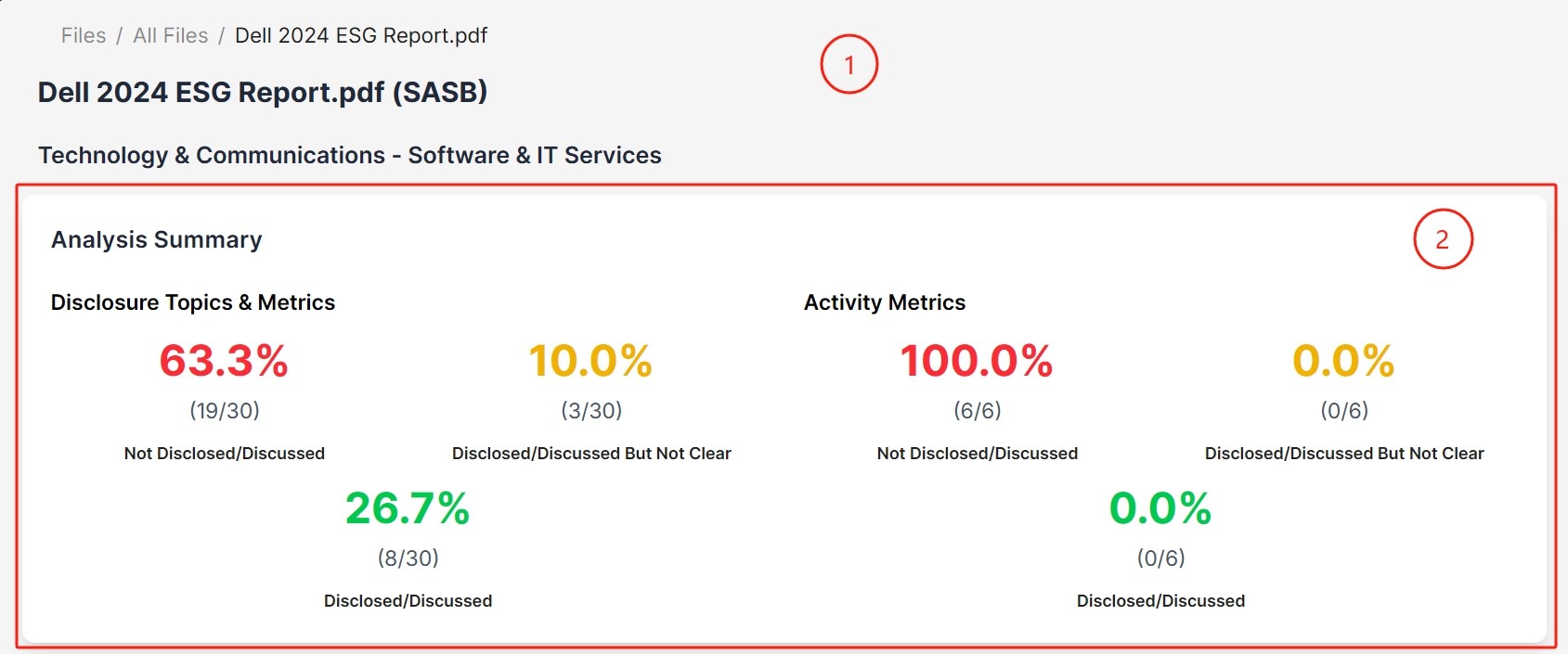}
  \hfill
  \includegraphics[width=0.35\textwidth]{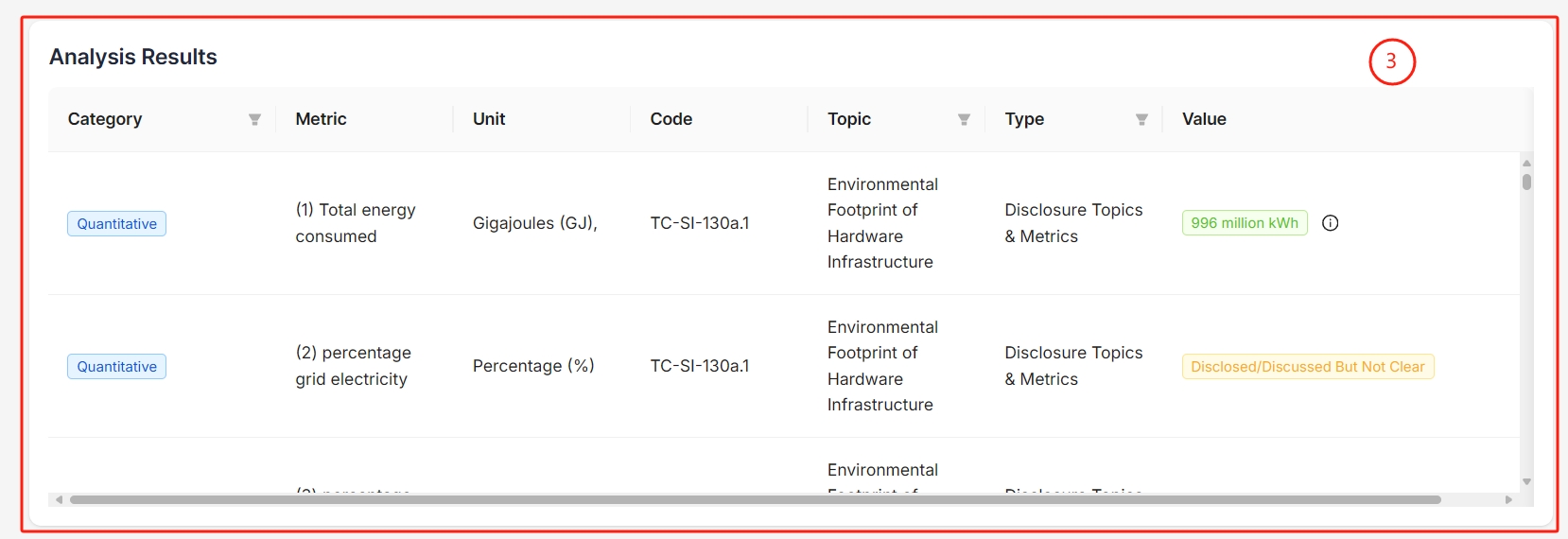}
  \hfill
  \includegraphics[width=0.32\textwidth]{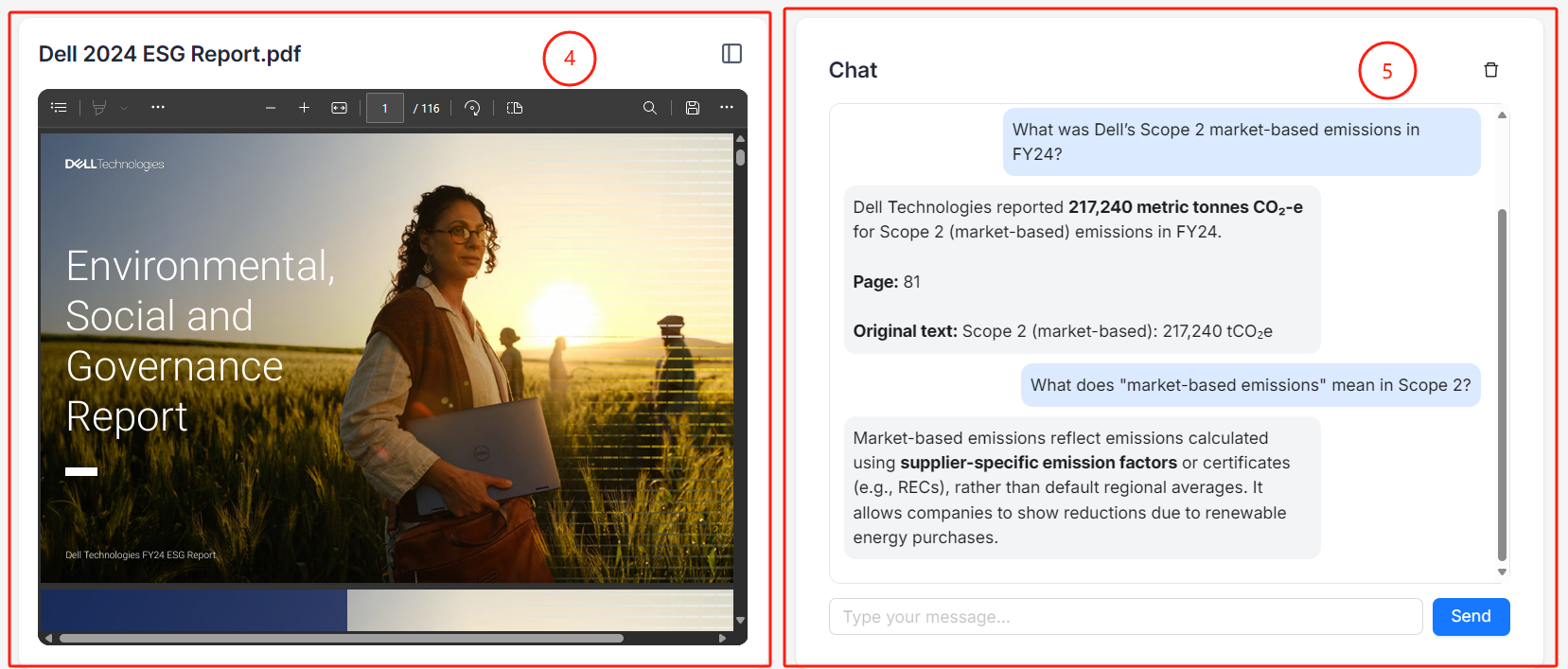}
  \vspace{-0.3cm}
  \caption{\euler interface showing key modules: Structured metric outputs, and PDF viewer, AI Chatbot.}
  \label{fig:esg-ui}
  
  \vspace{-0.4cm}
\end{figure*}

\vspace{-0.5cm}
\section{Performance Analysis}
\label{sec:experiments}

In this section, we evaluate \euler on a cross-industry case study based on the SASB standards. Our goals are to measure how well it populates this table from real ESG reports, together with the end-to-end runtime and API cost of the pipeline.

\stitle{Experimental Setup.}
We evaluate \euler\ on a SASB-aligned metric extraction task using four large, globally recognised companies, each representative of a different SASB industry: BMW, Macquarie Group, P\&G, and Dell. For each company, we take a single recent ESG report and process it end-to-end with \euler. Each company is paired with three manually selected SASB sub-industries, yielding 12 company-industry pairs in total. We assess \euler\ along two dimensions: (i) \emph{metric-level accuracy}, defined as the percentage of SASB metrics for which \euler's output (disclosure status and, when numeric, the value after unit normalisation) exactly matches the manually extracted ground truth, reported per pair as well as averaged at the company level and across all 12 pairs; and (ii) \emph{running time}, measured as the end-to-end wall-clock time from PDF upload to completion of the three sub-industry tables per company. To understand the impact of different LLM backends, we instantiate \euler with several recent models: Gemini-2.5-flash, Deepseek-V3.2-Exp, Claude-Haiku-4.5, GPT-5, and Qwen-3-32b.

% We select four large, globally recognised companies, each representative of a different SASB industry: (1)BMW (automotive manufacturing),  
% (2)Macquarie Group (diversified financial services),  
% (3)P\&G (consumer goods), and (4)Dell (hardware and related services). For each company, we use a single recent ESG or sustainability report in PDF form. These reports are processed end-to-end by \euler, including segmentation, retrieval over both standards and reports (dual-channel retriever), and LLM-based extraction and normalization to the SASB standard table. We pair each company with 3 sub-industries, thus, we have 12 company–sub-industry pairs in total.

% We evaluate \euler along two dimensions.  
% (1) \emph{Metric-level accuracy}: for each company–sub-industry pair, we compute the percentage of SASB metrics for which \euler's output (disclosure status and, when numeric, the value after unit normalisation) exactly matches the manually extracted ground truth. We report per-pair accuracy as well as averages at the company level and across all 12 pairs.  
% (2) \emph{Running time \& cost}: for each company, we measure end-to-end wall-clock time from PDF upload to completion of the three sub-industry tables and estimate the monetary cost of the run (LLM and embedding/API calls) based on public pricing of the underlying models.

\stitle{Results \& Discussion.} Across the 12 company-industry pairs, all models achieve high metric-level accuracy (Table~\ref{tab:model-per-pair}): Claude, DeepSeek, Qwen and Gemini reach overall averages of 0.93, 0.91, 0.93 and 0.94, respectively, while GPT-5 attains the highest average accuracy of 0.95. All models
are effectively perfect on the P\&G and Macquarie Group cases (company-level averages $\approx 1.0$), with most residual errors concentrated in the more heterogeneous BMW transportation and Dell
technology sub-industries. Runtime results reveal a clear efficiency-accuracy trade-off: Gtrade-off between efficiency and accuracy: GPT-5, while the st. Gemini-2.5-flash
and Qwen-3-32b offer a balanced middle ground, combining near-top accuracy with substantially lower latency than GPT-5, making them attractive choices for practical, interactive ESG analysis workloads.

\begin{table*}[t]
\centering

\begin{tabular}{l l l r r r r r}
\toprule
Company & Industry & Sub-industry & Claude & DeepSeek & GPT5 & QWen & Gemini \\
\midrule
\multirow{4}{*}{BMW}
  & \multirow{4}{*}{Transportation}
    & Auto Parts                    & 0.80 & 0.87 & 0.87 & 0.87 & 0.87 \\
  &                                   & Automobiles                  & 0.75 & 0.50 & 0.75 & 0.80 & 0.75 \\
  &                                   & Car Rental and Leasing    & 0.86 & 0.86 & 0.86 & 0.86 & 0.86 \\
  &                                   & \textit{Average}             & 0.80 & 0.74 & 0.82 & \textbf{0.84} & 0.82 \\
\midrule
\multirow{4}{*}{MCG}
  & \multirow{4}{*}{Financials}
    & Asset Manage Custody Activity & 0.93 & 1.00 & 1.00 & 1.00 & 1.00 \\
  &                                   & Commercial Banks            & 1.00 & 1.00 & 1.00 & 1.00 & 1.00 \\
  &                                   & Investment Banking Brokerage & 1.00 & 1.00 & 1.00 & 1.00 & 1.00 \\
  &                                   & \textit{Average}             & 0.98 & \textbf{1.00} & \textbf{1.00} & \textbf{1.00} & \textbf{1.00} \\
\midrule
\multirow{4}{*}{P\&G}
  & \multirow{4}{*}{Consumer Goods}
    & E-Commerce                    & 1.00 & 1.00 & 1.00 & 1.00 & 1.00 \\
  &                                   & Household Personal Products & 1.00 & 1.00 & 1.00 & 1.00 & 1.00 \\
  &                                   & Multiline and Specialty Retaile & 1.00 & 1.00 & 1.00 & 1.00 & 1.00 \\
  &                                   & \textit{Average}             & \textbf{1.00} & \textbf{1.00} & \textbf{1.00} & \textbf{1.00} & \textbf{1.00} \\
\midrule
\multirow{4}{*}{DELL}
  & \multirow{4}{*}{Technology \& Communications}
    & Hardware                      & 0.92 & 0.75 & 0.92 & 0.75 & 0.92 \\
  &                                   & Internet Media \& Services     & 1.00 & 1.00 & 1.00 & 0.96 & 0.96 \\
  &                                   & Telecommunication Services  & 0.93 & 0.93 & 0.97 & 0.97 & 0.97 \\
  &                                   & \textit{Average}             & 0.95 & 0.89 & \textbf{0.96} & 0.89 & 0.95 \\
\midrule
\multicolumn{2}{l}{\textbf{Overall Acc}}
  & \textit{Average}                & 0.93 & 0.91 & \textbf{0.95} & 0.93 & 0.94 \\
\midrule
\multicolumn{2}{l}{\textbf{Runtime(s)}}
  & \textit{Average}                & 275.37 & 459.32 & 1124.38 & 442.57 & 388.60 \\
\bottomrule
\end{tabular}
\caption{Company-industry metric-level accuracy and average runtime across different LLM backends.}
\label{tab:model-per-pair}
\vspace{-0.5cm}
\end{table*}

% \begin{table}[t]
% \centering

% \begin{tabular}{l r r r r r}
% \toprule
% Model & Runtime (s) \\
% \midrule
% Gemini-2.5-flash   & 388.60 \\
% Deepseek-V3.2-Exp  & 459.32 \\
% Claude-Haiku-4.5   & 275.37 \\
% GPT-5              & 1124.38 \\
% Qwen-3-32b         & 442.57 \\

% \bottomrule
% \end{tabular}
% \caption{Average End-to-end runtime(s).}
% \label{tab:model-runtime}
% \vspace{-2cm}
% \end{table}

\vspace{-0.5cm}
\section{Demonstration}
\label{sec:demo}

Figure~\ref{fig:esg-ui} illustrates the user interface of \euler, designed to support an interactive ESG analysis dashboard.  
% %
Key interface components include: 
\sstitle{(1) Report Header and Context}:  
    Displays the title of the uploaded ESG report, the applicable reporting framework (e.g., SASB), and the industry classification. 
\sstitle{(2) Disclosure Summary Panel}:  
    Provides a high-level overview of disclosure completeness. Metrics are grouped and color-coded according to the three categories defined earlier.
\sstitle{(3)Metrics Table}:  
    The core module presents extracted ESG data in a standardised, filterable table. Each row includes the metric dimensions (which may vary according to the standards) and the extracted value. 
    % \euler automatically classifies values based on the extracted text and the requirements defined by the applicable standard. 
    % It visually encodes the results using a consistent colour scheme, matching the disclosure status indicators shown in the discussion summary panel.
%
\sstitle{(4) PDF Viewer}:  
    Allows users to preview the uploaded report, enabling quick verification of extracted content and easy cross-referencing between metrics and their original textual context.
\sstitle{(5) ESG Chatbot}: 
    The integrated LLM-powered chatbot enhances user interaction by enabling natural language exploration of ESG reports. 
    It supports real-time clarification of terminology, dynamic querying of extracted content, and contextual understanding of disclosure topics.
    % and navigation assistance within the original document. 
    % This functionality makes ESG reports more accessible to users, enabling efficient exploration and rapid comprehension.

\stitle{User Scenarios.} 
\euler is designed as a versatile tool that serves users across the ESG ecosystem, each with distinct informational goals and decision-making contexts. 
% By automating the extraction, classification, and evaluation of ESG disclosures, the system lowers the technical barrier to accessing and interpreting corporate sustainability data.

\sstitle{Regulators, Investors, and Rating Providers}:  
    Use \euler to monitor ESG compliance, benchmark firms across sectors, and build consistent ESG scorecards. Scalable ingestion and standardised, traceable outputs support regulatory oversight, capital allocation, and third-party ratings on a common metric space.

\sstitle{Corporate Sustainability Teams and Consultants}:  
    Apply \euler for pre-publication self-assessment and multi-framework alignment. The system highlights missing or weakly substantiated metrics, streamlines ESG data collection and validation, and frees teams to focus on higher-level analysis and reporting strategy.

\sstitle{Civil Society and the General Public}:  
    Journalists, NGOs, advocacy groups, and individual stakeholders can rapidly extract and summarise ESG claims for fact-checking and thematic monitoring, while simplified visual summaries and metric-level drill-down make complex disclosures understandable to non-experts.

% By offering tailored assistance for a broad spectrum of users, from regulators and investors to corporate teams, \euler supports both institutional-scale oversight and public transparency. Its LLM-powered architecture enables scalable, context-aware interpretation of ESG disclosures, promoting more informed, equitable, and accountable decision-making across the sustainability landscape.

\section{Conclusion}
\label{sec:conclusion}
\euler presents a scalable and intelligent solution for automating the extraction, standardization, and evaluation of ESG reports. By leveraging LLM and a modular framework, it significantly reduces the manual effort required for ESG analysis and enables consistent benchmarking across industries and reporting frameworks. The system demonstrates strong potential in enhancing transparency, regulatory alignment, and decision-making for a diverse set of stakeholders. Its flexible architecture allows seamless adaptation to evolving ESG standards and domain-specific requirements.
Future work will focus on (1) expanding capabilities to process ESG disclosures in multiple languages, and (2) integrating with ESG ratings providers and financial data platforms.
% \begin{itemize}
%     \item \textbf{Multilingual Support:} Expanding capabilities to process ESG disclosures in multiple languages to accommodate global reporting practices.
%     \item \textbf{Extended Framework Coverage:} Incorporating additional ESG frameworks and taxonomies (e.g., ISSB, CSRD) to improve cross-standard interoperability and regulatory compliance.
%     \item \textbf{Ratings System Integration:} Connecting with ESG ratings providers and financial data platforms to enrich the interpretability and application of extracted metrics.
% \end{itemize}

% Ultimately, \euler aims to accelerate ESG transparency and accountability by transforming complex disclosures into actionable, auditable data at scale.

% \begin{acks}
% \end{acks}

\vspace{-0.2cm}
\bibliographystyle{ACM-Reference-Format}
\bibliography{ref}

@article{wang2023scalable,
  title={A Scalable Framework for Table of Contents Extraction from Complex ESG Annual Reports},
  author={Wang, Xinyu and Gui, Lin and He, Yulan},
  journal={arXiv preprint arXiv:2310.18073},
  year={2023}
}

@article{peng2024advanced,
  title={Advanced Unstructured Data Processing for ESG Reports: A Methodology for Structured Transformation and Enhanced Analysis},
  author={Peng, Jiahui and Gao, Jing and Tong, Xin and Guo, Jing and Yang, Hang and Qi, Jianchuan and Li, Ruiqiao and Li, Nan and Xu, Ming},
  journal={arXiv preprint arXiv:2401.02992},
  year={2024}
}

@article{shakil2021environmental,
  title={Environmental, social and governance performance and financial risk: Moderating role of ESG controversies and board gender diversity},
  author={Shakil, Mohammad Hassan},
  journal={Resources Policy},
  volume={72},
  pages={102144},
  year={2021},
  publisher={Elsevier}
}

@article{whelan2021esg,
  title={ESG and financial performance: Uncovering the relationship by aggregating evidence from 1,000 plus studies published between 2015--2020},
  author={Whelan, Tensie and Atz, Ulrich and Van Holt, Tracy and Clark, Casey},
  journal={New York: NYU STERN Center for sustainable business},
  pages={520--536},
  year={2021}
}

@article{friede2015esg,
  title={ESG and financial performance: aggregated evidence from more than 2000 empirical studies},
  author={Friede, Gunnar and Busch, Timo and Bassen, Alexander},
  journal={Journal of Sustainable Finance \& Investment},
  volume={5},
  number={4},
  pages={210--233},
  year={2015},
  publisher={Taylor \& Francis}
}

@article{khan2016corporate,
  title={Corporate sustainability: First evidence on materiality},
  author={Khan, Mozaffar N and Serafeim, George and Yoon, Aaron},
  journal={The Accounting Review},
  volume={91},
  number={6},
  pages={1697--1724},
  year={2016},
  publisher={American Accounting Association}
}

@misc{gri,
  author       = {{Global Reporting Initiative}},
  title        = {GRI Standards},
  year         = {2025},
  url          = {https://www.globalreporting.org/standards}
}

@misc{sasb,
  author       = {{Sustainability Accounting Standards Board}},
  title        = {SASB Standards},
  year         = {2025},
  url          = {https://sasb.ifrs.org/standards/}
}

@misc{tcfd,
  author       = {{Task Force on Climate-related Financial Disclosures}},
  title        = {Recommendations of the Task Force on Climate-related Financial Disclosures},
  year         = {2017},
  url          = {https://assets.bbhub.io/company/sites/60/2020/10/FINAL-2017-TCFD-Report-11052018.pdf}
}

@inproceedings{zhang2024mm,
  title={MM-LLMs: Recent Advances in MultiModal Large Language Models},
  author={Zhang, Duzhen and Yu, Yahan and Dong, Jiahua and Li, Chenxing and Su, Dan and Chu, Chenhui and Yu, Dong},
  booktitle={Findings of the Association for Computational Linguistics: ACL 2024},
  pages={1451--1466},
  year={2024},
  publisher={Association for Computational Linguistics}
}

@article{openai2023gpt4,
  title={GPT-4 Technical Report},
  author={OpenAI},
  journal={arXiv preprint arXiv:2303.08774},
  year={2023},
  url={https://arxiv.org/abs/2303.08774}
}

@article{rawat2024recent,
  title={Recent Advances in Generative AI and Large Language Models: Current Status, Challenges, and Perspectives},
  author={Rawat, Shubham and others},
  journal={arXiv preprint arXiv:2407.14962},
  year={2024}
}

@inproceedings{li2024citation,
  title={Citation-Enhanced Generation for LLM-based Chatbots},
  author={Li, Weitao and Li, Junkai and Ma, Weizhi and Liu, Yang},
  booktitle={Proceedings of the 62nd Annual Meeting of the Association for Computational Linguistics (Volume 1: Long Papers)},
  pages={1451--1466},
  year={2024},
  organization={Association for Computational Linguistics}
}

@article{chen2024bge,
  title={Bge m3-embedding: Multi-lingual, multi-functionality, multi-granularity text embeddings through self-knowledge distillation},
  author={Chen, Jianlv and Xiao, Shitao and Zhang, Peitian and Luo, Kun and Lian, Defu and Liu, Zheng},
  journal={arXiv preprint arXiv:2402.03216},
  year={2024}
}

@misc{baai2024bgereranker,
  author       = {Beijing Academy of Artificial Intelligence (BAAI)},
  title        = {BGE-Reranker-v2-M3},
  year         = {2024},
  howpublished = {\url{https://huggingface.co/BAAI/bge-reranker-v2-m3}},
  note         = {Multilingual reranking model by BAAI}
}

% \appendix

\end{document}